\documentclass[12pt,aps,showpacs,superscriptaddress,footinbib,preprint,noshowpacs]{revtex4-1}
\usepackage{color}
\usepackage{graphicx}
\usepackage{amsmath}
\usepackage{xcolor}
\usepackage{comment}
\usepackage{amssymb}
\usepackage[normalem]{ulem}

\usepackage{subfigure}

\begin{document}

\title{Bayesian semi-supervised learning for uncertainty-calibrated prediction of molecular properties and active learning}

\author{Yao Zhang}
\affiliation{Cavendish Laboratory, University of Cambridge, Cambridge CB3 0HE, United Kingdom }
\affiliation{Department of Applied Mathematics and Theoretical Physics, University of Cambridge, Cambridge CB3 0WA, United Kingdom}

\author{Alpha A. Lee}
\email{aal44@cam.ac.uk}
\affiliation{Cavendish Laboratory, University of Cambridge, Cambridge CB3 0HE, United Kingdom }

\begin{abstract}
Predicting bioactivity and physical properties of small molecules is a central challenge in drug discovery. Deep learning is becoming the method of choice but studies to date focus on mean accuracy as the main metric. However, to replace costly and mission-critical experiments by models, a high mean accuracy is not enough: Outliers can derail a discovery campaign, thus models need reliably predict when it will fail, even when the training data is biased; experiments are expensive, thus models need to be data-efficient and suggest informative training sets using active learning. We show that uncertainty quantification and active learning can be achieved by Bayesian semi-supervised graph convolutional neural networks. The Bayesian approach estimates uncertainty in a statistically principled way through sampling from the posterior distribution. Semi-supervised learning disentangles representation learning and regression, keeping uncertainty estimates accurate in the low data limit and allowing the model to start active learning from a small initial pool of training data. Our study highlights the promise of Bayesian deep learning for chemistry. 
\end{abstract}

\maketitle
\section{Introduction} 

Predicting physiological properties and bioactivity from molecular structure -- quantitative structure-property relationships (QSPR) -- underpins a large class of problems in drug discovery.  Classical QSPR workflows \cite{cherkasov2014qsar} separate descriptor generation -- mapping a 2D  \cite{randic1991generalized,ivanciuc2000qsar,durant2002reoptimization,rogers2005using} or 3D molecular structure \cite{cramer1988comparative,verma20103d} into a vector of real numbers using some handcrafted rules -- and machine learning method that connects descriptors to property. Pioneering advances in machine learning such as graph neural networks directly take a molecular graph as input and infer the optimal structure-to-descriptor map from data \cite{scarselli2009graph,duvenaud2015convolutional}, outperforming classical machine learning methodologies with handcrafted descriptors \cite{wu2018moleculenet}. 

Nonetheless, graph neural networks are usually developed using frequentist maximum likelihood inference, with the benchmark being the mean error on a test set. However, if the goal of QSPR is to replace mission-critical but expensive experiments, a low mean error is insufficient: the user needs to have an estimate of uncertainty and know when the model is expected to fail. This is because typically only a small number of top-ranked predictions are selected to test experimentally, thus outliers can ruin a discovery campaign. Moreover, cost limits the number of experiments that can be run, thus an approach that judiciously designs the training set to maximise information gained is needed.   

Uncertainty quantification, or domain applicability, has been extensively considered in the QSPR literature but not in the context of graph neural networks and not in a statistically complete way. Previous works estimate uncertainty of prediction as the distance in descriptor space between the input molecule and the training set, or training an ensemble of models and evaluating the variance \cite{sheridan2004similarity,sushko2010applicability,sheridan2012three, toplak2014assessment}. More recent works consider conformal regression \cite{norinder2014introducing,svensson2017modelling}, which trains two  models, one for the molecular property and one for the error. However, there are two sources of uncertainty: Epistemic uncertainty arises due to insufficient data in the region of chemical space that the model is asked to make predictions on. Aleatoric uncertainty arises due to noise in the measurements themselves (e.g. noisy biochemical assays) \cite{kendall2017uncertainties}. Distance to the training set and variance within a model ensemble approximately capture epistemic uncertainty, whilst employing an ancillary model for prediction error approximately captures aleatoric uncertainty. We will show that the Bayesian statistical framework captures both sources of uncertainty in a unified and statistically principled manner. 

Active learning strategies have been considered in the drug discovery literature \cite{reker2015active,reker2017active}. However, those pioneering works considered a priori defined molecular descriptors, and estimate uncertainty via variance within an ensemble of models. Notwithstanding the shortcomings with incomplete modelling of uncertainty discussed above, employing graph neural networks in active learning presents unique opportunities and challenges: High model accuracy in the big data limit comes at the cost of being data-hungry. As the descriptor is fully data-driven, the model cannot estimate how ``far'' a compound is from the test set in the low-data limit, leading to poor uncertainty estimate and breaking down the active learning cycle. Low-data drug discovery has been considered in the context of one-shot learning \cite{altae2017low} which estimates distance in chemical space by pulling data from related tasks. Nonetheless, this approach requires a priori knowledge on which tasks are related. Works on generative molecular design overcome this problem \cite{gomez2018automatic} by starting the active learning cycle with $<1000$ quantitative measurements, which impose a significant upfront experimental cost. 

In this paper, we combine Bayesian statistics -- a principled framework for uncertainty estimation --  with semi-supervised learning which learns the representation from unlabelled data. We show that Bayesian semi-supervised graph convolutional neural networks can robustly estimate uncertainty even in the low data limit and drive an active learning cycle, and overcome dataset bias in the training set. Further, we demonstrate that the quality of posterior sampling is directly related to accuracy of the uncertainty estimates. As different Bayesian inference methods can be mixed and matched with different models, our study opens up a new dimension in the design space of uncertainty-calibrated QSPR models.

\section{Methods and Data} 
A machine learning method has two independent components: model and inference. The model is a function with parameters that relate the input to the output. Inference pertains to the methodology by which the model parameters are inferred from data. In terms of model, we focus on graph convolutional neural network models that take molecular graphs as input. In terms of inference, we focus on the Bayesian methodology. 

\subsection{Supervised graph convolutional neural network} 
Our baseline model is the graph convolutional fingerprint model \cite{duvenaud2015convolutional}. The salient idea is the message passing operation \cite{gilmer2017neural}, which creates a vector that summarises the local atomic environment around each atom while respecting invariance with respect to atom relabelling. A molecule is described by a graph, where the nodes are atoms and the edges are bonds. Atom $i$ is described by a vector of atomic properties $\mathbf{x}_{v} $, and a bond connecting $i$ and $j$ is described by bond properties $\mathbf{e}_{vw}$. The algorithm is iterative: at step $t$, each atom has a hidden state $\mathbf{h}_{v}^{t}$, which depends on ``messages'' $\mathbf{m}_{v}^{t}$ received from surrounding atoms as well as $\mathbf{h}_{v}^{t-1}$. The hidden states can be interpreted as descriptors of local atomic environment, and the messages allow adjacent atoms to comprehend the environment of its neighbours. Each atom is initialised to its atomic features, $\mathbf{h}_{v}^0 = \mathbf{x}_{v}$, and 
\begin{equation}
\mathbf{m}_{v}^{t} = \sum_{w \in \mathcal{N}(\mathbf{v})} (\mathbf{h}_{w}^{t-1}, \mathbf{e}_{vw}), 
\label{message}
\end{equation}   
and 
\begin{equation}
\mathbf{h}_{v}^{t} = \sigma(\mathbf{H}_{t-1}^{\mathrm{deg}(v)} \mathbf{m}_v^{t}),
\label{update} 
\end{equation}   
where $\mathcal{N}(\mathbf{v})$ denotes the set of atoms bonded to atom $v$, $\sigma(\cdot)$ is the sigmoid function,  $\mathbf{H}_t^{N}$ is a learned matrix for each step $t$ and vertex degree $N$. The algorithm is run $T$ times, with $T$ being a hyperparameter. In the final step, the output is given by a multilayer neural network $f(\cdot)$ that takes a weighted average of the hidden states at each step as input return a prediction, 
\begin{equation}
y = f\left(\sum_{v,t} \mathrm{softmax}(\mathbf{W}_t \mathbf{h}_{v}^{t} ) \right)
\label{aggregation}
\end{equation}
where $\mathbf{W}_t$ are learned readout matrices, one for each step $t$.

We use the implementation reported in the repository \footnote{$\mathtt{https://github.com/debbiemarkslab/neural-fingerprint-theano}$}.  In all experiments, we consider $T=3$, hidden layer at each level has 128 units, fingerprint length 256 (i.e. $\mathbf{W}_t \in \mathbb{R}^{128 \times 256}$), and $f(\cdot)$ is a two layer neural network with 128 units each and $relu$ units.

\subsection{Semi-supervised graph convolutional neural network} 
The fully supervised approach learns molecular descriptors directly from data. This is an advantage if one has a lot of data but a disadvantage in the data-limited settings such as active learning applications, where the objective is to design informative experiments starting from a small pool of initial training data. 

The insight behind the semi-supervised approach is that significant amount of chemical knowledge is contained within the molecular structures themselves, without any associated molecular properties (i.e. unlabelled data). Thermodynamic stability puts constrains on what bonds are possible, and tends to put certain bonds near each other, forming persistent chemical motifs. For example, just by looking at drug molecules, one would immediately spot ubiquitous motifs such as amide group, benzene rings etc., and some motifs often occur together as scaffolds \cite{welsch2010privileged}. The key assumption is that those persistent chemical motifs contribute to the molecular property that we want to predict. We can  make mathematical progress by constructing a descriptor akin to Equation (\ref{message}) - (\ref{aggregation}). However, the objective is no longer trying to fit a particular property. Rather, the hidden states $\mathbf{h}_{v}^{t}$, which summarises the atomic environment around atom $v$ within radius $t$, are constructed such that they are predictable from the hidden states of the surrounding atoms. Therefore, the model learns a descriptor that clusters similar environments. 

Specifically, we use the semi-supervised approach developed by Ngyen et al. \cite{nguyen2017semi}, which builds on the paragraph vector approach in natural language processing \cite{le2014distributed}. Given a set of molecular structures $\mathcal{M}$, the hidden states $\mathbf{h}_{v}^{t}$ maximise the log-likelihood 
\begin{equation}
L =  \sum_{m \in \mathcal{M}} \sum_{v \in m} \sum_{t=1}^{T} \log P(\mathbf{h}_{v}^{t}| \mathbf{u}_m), 
\label{objective}
\end{equation} 
\begin{equation}
P(\mathbf{h}_{v}^{t}| \mathbf{u}_m) = \frac{\exp\left( (\mathbf{h}_{v}^{t})^{T} \mathbf{u}_m \right) }{\sum_{n \in \mathcal{M}}   \exp\left( (\mathbf{h}_{v}^{t})^{T} \mathbf{u}_n \right) }, 
\end{equation}
where $\mathbf{u}_n$ is the molecular identifier, obtained by maximising Equation (\ref{objective}), with $\mathbf{h}_{v}^{t}$ defined by Equation (\ref{message})-(\ref{update}). We can interpret $\mathbf{u}_n$ as a vector that describes the ``type'' of molecule, and the objective encourages the hidden states $\mathbf{h}_{v}^{t}$ to take values such that similar molecules have similar atomic environments. 

After finding parameters that maximise the objective (\ref{objective}), $\{\mathbf{h}_{v}^{t}\}$ are then passed to a neural network, Equation (\ref{aggregation}). The parameters of the neural network as well as the readout matrices $\mathbf{W}_t$ are learned in a supervised manner. Note that this formalism infers descriptors using unsupervised learning and uses supervised learning to relate descriptors to molecular properties. 

We use the implementation reported in the Github repository\footnote{$\mathtt{https://github.com/pfnet-research/hierarchical-molecular-learning}$} accompanying ref \cite{nguyen2017semi}.  In all experiments, we consider $T=3$, hidden layer at each level has 128 units, fingerprint length 256 (i.e. $\mathbf{W}_t \in \mathbb{R}^{128 \times 256}$), and $f(\cdot)$ is a two layer neural network with 128 units each and $relu$ units.
 
\subsection{Bayesian deep learning} 
In Bayesian inference, the aim is to determine the distribution of model parameters that conforms to the data, the so-called posterior distribution. Let $\boldsymbol{\theta}$ be model parameters, $\mathbf{x}_i$ the dependent variables and $y_i$ the independent variable, such that 
\begin{equation}
y_i = F(\mathbf{x}_i,\boldsymbol{\theta}) + \epsilon_i .
\label{model}
\end{equation} 
where $\epsilon_i \sim \mathcal{N}(0,\sigma^2_i)$ is the measurement noise.  Bayes theorem states the posterior distribution, $P(\boldsymbol{\theta}|\{\mathbf{x}_i\},\{y_i\})$, is related to the likelihood, $P(\{y_i\} |\boldsymbol{\theta},\{\mathbf{x}_i\})$ and the prior $P(\boldsymbol{\theta})$ via 
\begin{equation}
P(\boldsymbol{\theta}|\{\mathbf{x}_i\},\{y_i\}) = \frac{1}{Z} P(\{y_i\} |\boldsymbol{\theta},\{\mathbf{x}_i\}) P(\boldsymbol{\theta}), 
\end{equation}  
where $Z$ is a normalising constant. The prediction for an unknown input $\mathbf{\hat{x}}$ is obtained by averaging over the posterior 
\begin{equation}
\left< \hat{y}\right> = \int P(\boldsymbol{\theta}|\{\mathbf{x}_i\},\{y_i\}) F(\mathbf{\hat{x}},\boldsymbol{\theta}) \;  \mathrm{d}\boldsymbol{\theta}. 
\label{mean}
\end{equation}
The uncertainty of model predictions can be readily derived from this Bayesian formalism. There are two types of uncertainties \cite{kendall2017uncertainties}. First, the epistemic uncertainty, is given by the variance of the prediction with respect to the posterior 
\begin{equation}
\mathrm{var}(\hat{y}) = \int P(\boldsymbol{\theta}|\{\mathbf{x}_i\},\{y_i\}) ( \left< \hat{y}\right> - F(\mathbf{\hat{x}},\boldsymbol{\theta}))^2 \;  \mathrm{d}\boldsymbol{\theta}.  
\label{variance}
\end{equation}
Second, the aleatoric uncertainty, is the intrinsic noise of the measurement $\sigma_i^2$. This aleatoric noise can depend on the input, $\sigma_i^2 = \sigma(\mathbf{x})^2$, as certain areas of the chemical space can be intrinsically more variable. 

We note that the log posterior is, up to a constant,  
\begin{equation} 
-\log P(\boldsymbol{\theta}|\{\mathbf{x}_i\},\{y_i\}) =  \sum_{i} \left(\frac{1}{2 \sigma_i^2} (y_i -F(\mathbf{\hat{x}},\boldsymbol{\theta}))^2  + \frac{1}{2} \log \sigma_i^2 \right) - \log P(\boldsymbol{\theta}), 
\label{loss_fn}
\end{equation} 
which is exactly the mean-squared loss if $\sigma_i$ is constant, with $\log P(\boldsymbol{\theta})$ being the regulariser. Therefore, maximum likelihood inference is a special case of Bayesian inference. 

The Bayesian formalism is easy to state but computationally expensive. The numerical bottleneck is the numerical evaluation of the high dimensional integrals (\ref{mean})-(\ref{variance}).  A plethora of approximate numerical methods have been developed in the literature to overcome this bottleneck. However, there is no free lunch, and methods which approximate the posterior well are usually computationally expensive. In this paper, we will consider two approximate methods spanning the cost-accuracy spectrum. 

\subsubsection{Dropout variational inference}
Variational inference seeks to approximate the posterior distribution by a distribution that is much easier to sample from. Refs \cite{gal2015dropout,kendall2017uncertainties} show that a popular way to regularise neural networks -- dropout -- is equivalent to approximate Bayesian inference. The algorithm is simple: The neural network is forked at the last layer to have two outputs, the predicted aleatoric uncertainty $\sigma_i^2$ and dependent variable $y_i$, and trained to minimise the loss (\ref{loss_fn}). However, during training, each unit has a probability $p$ of being set to 0.

For a neural network with $M$ units, refs \cite{gal2015dropout,kendall2017uncertainties} show that the above algorithm is approximately equal to finding parameters $\boldsymbol{\theta} = (\boldsymbol{\Theta_1}, \boldsymbol{\Theta}_2 \cdots \boldsymbol{\Theta}_M)$ that fit the distribution  
\begin{equation}
q(\boldsymbol{\theta}) = \prod_{m=1}^M  \boldsymbol{\Theta_m} z_m , \;\;  z_m \sim \mathrm{Bernoulli}(p), 
\label{trial_dist}
\end{equation}
to the posterior distribution $P(\boldsymbol{\theta}|\{\mathbf{x}_i\},\{y_i\})$, where $\Theta_m$ is the parameter vector associated with the $m^{th}$ unit.  

Distribution (\ref{trial_dist}), although not the same as the true posterior distribution, is significantly easier to sample: In the prediction phase, the model is run $N$ times, and akin to the training phase each unit has probability $p$ of being set to 0. The final prediction and total uncertainty is taken to be the mean over $N$ different predictions of depending variable and variance, $\{y^i, (\sigma^{i})^2\}_{i=1}^{N}$, 
\begin{equation}
\left<y\right> = \frac{1}{N}\sum_{m=1}^{N} y^{m}, \quad \mathrm{var}(y) =\frac{1}{N} \sum_{m=1}^{N} (y^{m} -\left<y\right>)^2 + \frac{1}{N} \sum_{m=1}^{N}  (\sigma^{m})^2 .  
\label{total_variance}
\end{equation}
The first term in Equation (\ref{total_variance}) is the epistemic uncertainty and the second term is the aleatoric uncertainty. 

In our numerical experiments, dropout is applied to every unit that is trained using supervised learning, i.e. every unit in the supervised graph convolutional neural network is subjected to dropout, whereas for the semisupervised case the layers on top of the hidden states are trained with dropout.

\subsubsection{Stein Variational Gradient Descent}

Rather than fitting a distribution to the posterior, Stein Variational Gradient Descent (SVGD) \cite{liu2016stein} directly draws samples from the posterior via gradient descent. Specifically, let $\{ \boldsymbol{\theta}^{0}_i\}_{i=1}^{N}$ be parameters randomly and independently initialised in parameter space. We want to evolve parameters such that, after $T$ steps, $\{ \boldsymbol{\theta}^{T}_i\}_{i=1}^{N} $ are $N$ independent samples drawn from $P(\boldsymbol{\theta}^{t}_j|\{\mathbf{x}_i\},\{y_i\})$.  Ref \cite{liu2016stein} shows that the following dynamical system does the trick:  
\begin{equation}
\boldsymbol{\theta}^{t+1}_i  = \boldsymbol{\theta}^{t}_i  + \eta \phi(\boldsymbol{\theta}^{t}_i), 
\label{gradient_des}
\end{equation} 
where 
\begin{equation}
\phi(\boldsymbol{\theta}) = \frac{1}{N} \sum_{j=1}^{N}  \left[ k(\boldsymbol{\theta}^{t}_j,\boldsymbol{\theta}) \nabla_{\boldsymbol{\theta}^{t}_j} \log P(\boldsymbol{\theta}^{t}_j|\{\mathbf{x}_i\},\{y_i\}) +  \nabla_{\boldsymbol{\theta}^{t}_j}  k(\boldsymbol{\theta}^{t}_j,\boldsymbol{\theta}) \right]. 
\label{forces}
\end{equation} 
and $ k(\cdot,\cdot) $ is a generic kernel function and $\eta$ is the learning rate. Equations (\ref{gradient_des})-(\ref{forces}) can be interpreted as free energy minimisation of an interacting particle system: a ``particle'' (parameter vector) is subjected to a ``force'' $\phi(\boldsymbol{\theta})$, which drives particles to regions of low energy (low loss), whilst forcing the particles apart to maximise entropy. The total uncertainty is evaluated also with Equation (\ref{total_variance}), except $\{y^{m}\}_{m=1}^{N}$ are predictions from different model parameters $\{ \boldsymbol{\theta}_i\}_{i=1}^{N} $.

The key advantage of Equation (\ref{gradient_des}) - (\ref{forces}) is that it is a well-defined approximation: Frequentist inference (c.f. Equation (\ref{loss_fn})) is recovered if $N = 1$, whereas when $N \rightarrow \infty$ the system exactly samples from the posterior. Therefore, for finite $N$, the algorithm interpolates between frequentist and full Bayesian inference. The computational cost and memory demands increase with $N$, and in this paper we use $N=50$.  

\begin{figure}[!h]
\centering
\includegraphics[scale=0.20]{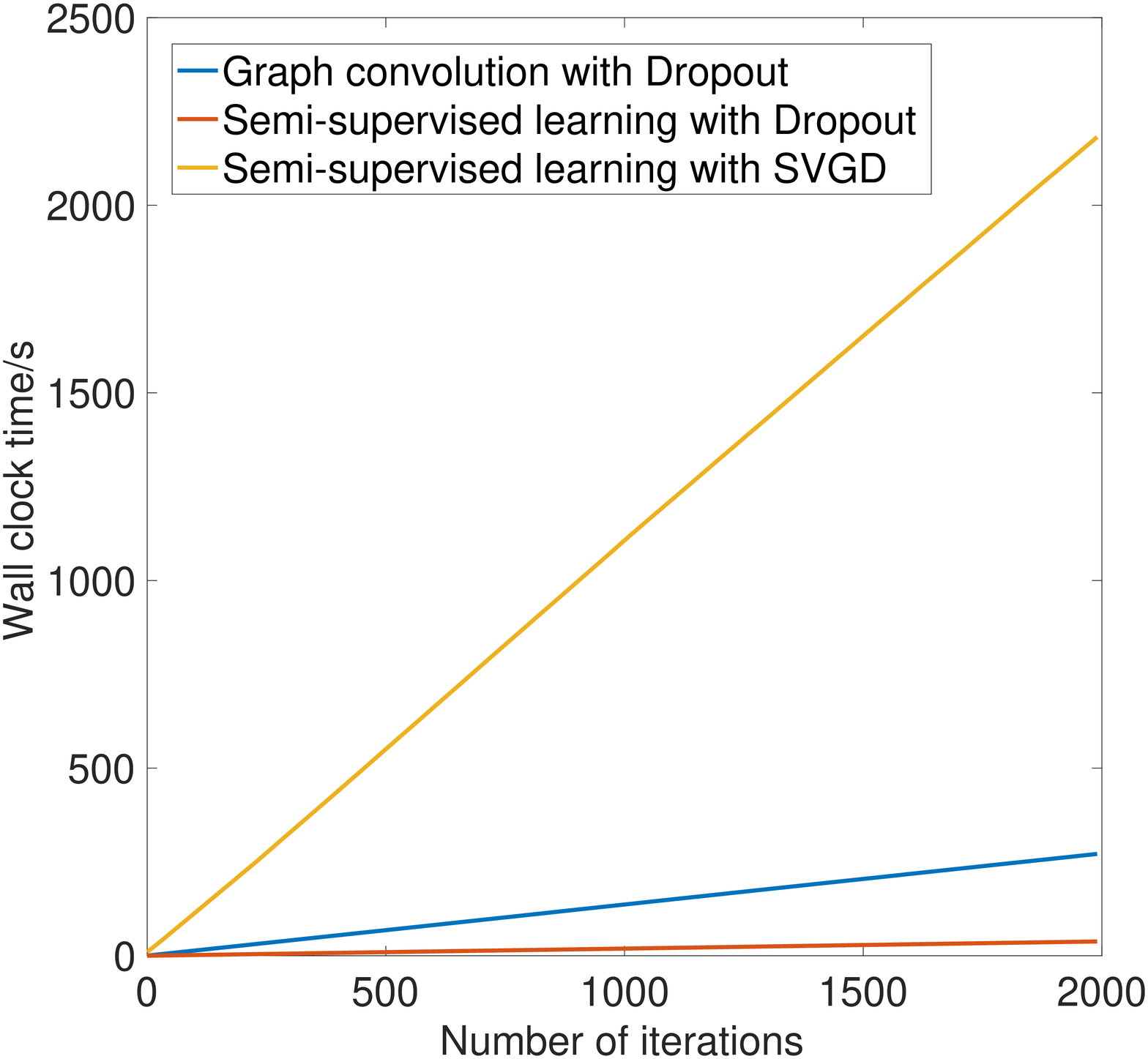}
\caption{The wall clock time on a Nvidia P100 GPU of graph convolution with dropout, semi-supervised with dropout, and semi-supervised with Stein Variational Gradient Descent on the melting point dataset.}
\label{wall_clock}
\end{figure}

To illustrate the computational demands of Stein Variational Gradient Descent, Figure \ref{wall_clock} shows the wall clock time, on a Nvidia P100 GPU, as a function of the number of gradient updates steps for graph convolution with dropout, semi-supervised with dropout, and semi-supervised with SVGD on the melting point dataset discussed below. Both SVGD and Stochastic Gradient Descent use back-propagation to optimize the neural network parameters. For models trained using Stochastic Gradient Descent, the computational complexity of back-propagation at each iteration is $O(BM)$, where $B$ is the number of training samples at each iteration and $M$ is the number of parameters in the model. The semi-supervised model has less parameters than the fully supervised model, thus the wall-clock time is less per iteration. In SVGD, we need to update $N$ Stein particles per iteration, thus wall clock time per iteration scales as $O(BMN)$. 

\subsubsection{Architectures and hyperparameters}
As the objective of this paper is to demonstrate the types of chemical problems that Bayesian deep learning can tackle, we adopt common parameters for graph convolutional neural networks taken from the literature rather than performing extensive hyperparameter optimisation and neural architecture search. For both supervised and semi-supervised graph convolutional neural networks, we keep the number of hidden layers the same as the original implementation in the GitHub repositories cited above. Following the implementations in the repositories we keep the dimension of the fingerprint twice of $n_h$, the number of neurons in the hidden layers, and only optimise the $n_h$ by a grid search over $\{32, 64,128,256\}$, choosing the value of $n_h$ with the best averaged 5-fold cross-validation root mean squared error over all the datasets. This leads to the architecture of $T = 3$, $N = 50$, hidden units = 128 and two layers. 

\subsection{Datasets} 
We consider a set of common regression benchmarks for physical properties prediction and bioactivity prediction. The $\mathtt{Melting Point}$ dataset is a collection of 3,025 melting point measurements of drug-like molecules used in a benchmark study \cite{coley2017convolutional}. The $\mathtt{ESOL}$ dataset is a set of 1,128 measured aqueous solubilities of organic molecules \cite{delaney2004esol}, and the $\mathtt{FreeSolv}$ dataset is a set of 643 hydration free energy measurements small molecules in water \cite{mobley2014freesolv}. The $\mathtt{ESOL}$ and $\mathtt{FreeSolv}$ datasets are used in the MoleculeNet benchmark \cite{wu2018moleculenet}. The $\mathtt{CatS}$ dataset comprises half-maximal inhibitory concentration $\log \mathrm{IC}_{50}$ measurements of 595 molecules against Cathepsin S, taken from the D3R Grand Challenge 3 and 4 \cite{Gaieb2019}. The $\mathtt{Malaria}$ dataset is a set of in vitro half-maximal effective concentration ($\log \mathrm{EC}_{50}$) measurement of 13,417 molecules against a sulfide-resistant strain of P. falciparum, the parasite that causes malaria \cite{gamo2010thousands} used in benchmark \cite{duvenaud2015convolutional}. The $\mathtt{p450}$ dataset is a dataset of half-maximal effective concentration measurements of 8,817 molecules against Cytochrome P450 3A4, a key enzyme for metabolism and clearance of xenobiotics, taken from the PubChem assay AID 1851.  

To give the reader a sense of how ``hard'' the different datasets are, we consider a simple baseline model of XGBoost \cite{chen2016xgboost} on ECFP6 fingerprints \cite{rogers2010}. We split the data into 80/10/10 (training/validation/testing). We take $\mathrm{MaxDepth}=5$, $\mathrm{LearningRate} = 0.01$, and optimise the number of estimators ($\mathrm{nEstimators} = [50,100,150,200,250]$) using the validation set. Table \ref{baseline_model} shows the coefficient of determination $R^2$ and root mean squared error (RMSE). Judging from the coefficient of determination, the dataset difficulty is (from easiest to hardest) $\mathtt{FreeSolv}<\mathtt{Melting}<\mathtt{ESOL}<\mathtt{CatS}<\mathtt{Malaria}<\mathtt{p450}$.

\begin{table}[!h]
\centering
\begin{tabular}{|l |l |l|l|}
\hline
\textbf{Dataset} & \bf{Number of data points}  &  $\mathbf{R^2}$ &  \textbf{RMSE} \\ \hline
$\mathtt{FreeSolv}$       &    643   & 0.765      & 1.90          \\ \hline
$\mathtt{ESOL}$            &   1128  & 0.704      & 1.14         \\ \hline
$\mathtt{CatS}$              &   595   & 0.654      & 0.367         \\ \hline
$\mathtt{Melting Point}$ &   3025  &  0.725       &50.2          \\ \hline
$\mathtt{p450}$              &   8817 & 0.291       & 0.996         \\ \hline
$\mathtt{Malaria}$          &   13417 & 0.499       & 0.655        \\ \hline
\end{tabular}
\caption{The performance of the baseline XGBoost model on the datasets considered in this paper.}
\label{baseline_model}
\end{table}

\section{Results}

\subsection{Uncertainty quantification} 
We first consider how well can the model estimate its own uncertainty given the full dataset, split into training (80\%) and test (20 \%) sets. The quality of the uncertainty estimate is operationalised by asking what is the model accuracy when the most uncertain predictions are removed, with uncertainty quantified by the variance computed from Equation (\ref{total_variance}). Our baseline method is graph convolution with dropout, which has recently been implemented in the DeepChem package as a feature \cite{wu2018moleculenet}, although to our knowledge this is the first study that benchmarks Bayesian graph convolutional neural networks in terms of uncertainty quantification. 

\begin{figure*}[!h]
\centering
\includegraphics[scale=0.25]{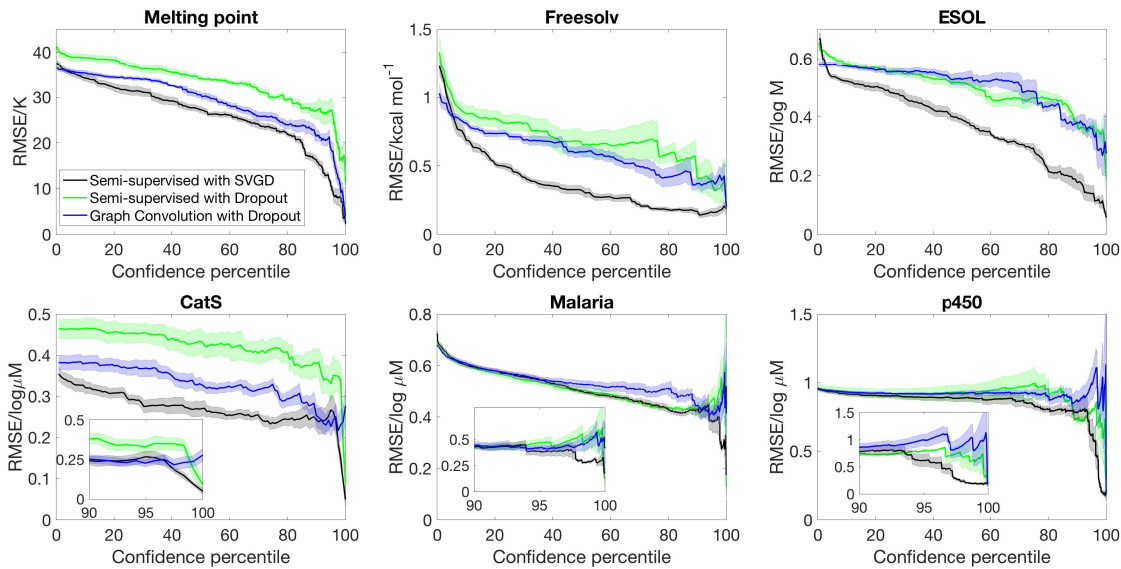}
\caption{Bayesian semi-supervised learning accurately predicts molecular properties and uncertainty. The plots show the model accuracy on the test set as a function of confidence percentile. The inset highlights the fact that predictions which the model are confident about are indeed accurate.  }
\label{uncertainty} 
\end{figure*}

Figure \ref{uncertainty} shows that semi-supervised learning with Stein Variational Gradient Descent accurately estimates uncertainty and significantly outperforms the baseline on every dataset. The plots show how the test set error varies as a function of confidence percentile -- i.e. what is the error if we only consider the top-$n$ \% of compounds in the test set ranked by confidence (note that confidence is inverse of the uncertainty, quantified by Equation (\ref{total_variance})); the shaded region is one standard deviation, estimated by analysing 5 random partitions of the data into training and test sets. In every case, the error is a decreasing function of model confidence, thus the model successfully estimates which predictions are likely to be correct and which predictions are outliers. 

Another metric that we can evaluate is the shape of the confidence-error curve. For $\mathtt{ESOL}$ and $\mathtt{FreeSolv}$, the error is a steeply decreasing at the low confidence limit before plateauing, suggesting that most predictions are accurate but for a few outliers, which the Bayesian method can identify. The situation is different for $\mathtt{Melting Point}$, $\mathtt{Malaria}$ and  $\mathtt{p450}$ -- the error is slowly decreasing at the low confidence limit before sharply decreasing when it approaches the 100\% confidence percentile limit (see also insets of Figure \ref{uncertainty}). This suggests that a few predictions are very accurate, and Bayesian method can pick out those accurate predictions amid many less accurate ones. We note that our Bayesian model is well-suited for virtual screening applications, where the challenge is ensuring that the top-ranked actives picked out by the model are indeed actives, since only a very small proportion of the compounds ranked will actually be screened experimentally (the "early recognition problem") \cite{bender2005discussion,truchon2007evaluating}. 

A lingering question whether the quality of the uncertainty estimate is due to a set of good descriptors (obtained via semi-supervised learning) or accuracy of the Bayesian methodology. Figure \ref{uncertainty} also shows that replacing Stein Variational Gradient Descent with dropout significantly reduces model performance. At the same confidence percentile, Stein Variational Gradient Descent consistently outperforms dropout. This suggests that the quality of posterior sampling drastically impacts the quality of uncertainty estimation. 

The quality of uncertainty estimates can also be gauged by the correlation between the predicted uncertainty on test data points and the error that the model incurs. Table \ref{rank_corr} shows the Spearman correlation coefficient between the predicted variance and model error. As expected, combining semi-supervised learning with Stein Variational Gradient Descent leads to method with the highest rank correlation. This result is consistent with Figure \ref{uncertainty}, which shows that semi-supervised learning with Stein Variational Gradient Descent has the lowest confidence-error curve.  Moreover, the rank correlation between predicted uncertainty and error broadly (although not exactly) follows the ``difficulty'' of the data (c.f. Table \ref{baseline_model}) -- $\mathtt{FreeSolv}$ and $\mathtt{ESOL}$ have the highest rank correlation, and $\mathtt{p450}$ has the lowest.  

\begin{table}
\centering
\begin{tabular}{|l|l|l|l|}
\hline
 \textbf{Dataset} & \textbf{\begin{tabular}[c]{@{}l@{}}Graph convolution \\ with Dropout\end{tabular}} &  \textbf{\begin{tabular}[c]{@{}l@{}}Semi-supervised \\ with Dropout\end{tabular}}  & \textbf{\begin{tabular}[c]{@{}l@{}}Semi-supervised \\ with SVGD\end{tabular}} \\ \hline
$\mathtt{FreeSolv}$         &  $0.531\pm 0.061$                                &$0.439 \pm 0.093$                            & $\mathbf{0.688 \pm 0.053}$                             \\ \hline
$\mathtt{ESOL}$              &  $0.112 \pm 0.035$                                & $0.306\pm 0.079$                           &$\mathbf{0.553 \pm 0.026}$                              \\ \hline
$\mathtt{CatS}$                &  $0.049 \pm 0.036$                                & $0.066\pm 0.044$                          & $\mathbf{0.310 \pm 0.019}$                            \\ \hline
$\mathtt{Melting Point}$    & $0.192 \pm 0.016$                                  & $0.284 \pm 0.035$                       & $\mathbf{0.337 \pm 0.013}$                            \\ \hline
$\mathtt{p450}$             &  $0.167 \pm 0.015$                                  & $0.185 \pm 0.049$                          & $\mathbf{0.213\pm 0.010}$                              \\ \hline
$\mathtt{Malaria}$          &$0.315 \pm 0.028$                                  &$0.317 \pm 0.031$                                 & $\mathbf{0.378 \pm 0.019}$                              \\ \hline
\end{tabular}
\caption{Combining semi-supervised learning with with SVGD yields uncertainty estimates that are strongly correlated with actual model error. The table shows the Spearman correlation coefficient between the variance predicted by the model and the error on test data points.}
\label{rank_corr}
\end{table}

\begin{figure*}[!h]
\centering
\includegraphics[scale=0.22]{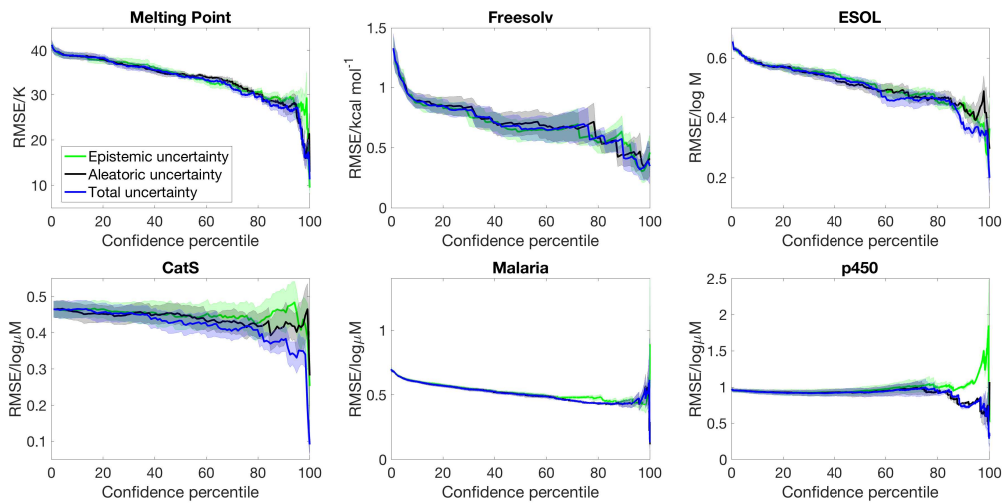}
\caption{Epistemic uncertainty and aleatoric uncertainty are distinct sources of uncertainty, and a combination of them is needed to obtain a good estimate of model error. We plot the confidence-error curve for semi-supervised learning with Stein Variational Gradient Descent where the confidence is estimated from combining epistemic and aleatoric uncertainty, epistemic uncertainty alone, and aleatoric uncertainty alone. }
\label{types_of_uncertainty} 
\end{figure*}  

Previous works on domain applicability have focused on either building an auxiliary model to predict the error \cite{norinder2014introducing,svensson2017modelling}, or estimating the uncertainty of a prediction via the distance of the input to the training set \cite{sheridan2004similarity,sushko2010applicability, toplak2014assessment}. The former models aleatoric uncertainty whereas the latter approximately captures epistemic uncertainty. Our Bayesian method captures both sources of uncertainty in a statistically principled manner. However, our model also provides independent estimates of epistemic and aleatoric uncertainties. As such, we can ask the question: is knowing epistemic or aleatoric uncertainty alone sufficient to estimate whether a prediction is accurate? 

Figure \ref{types_of_uncertainty} shows that the confidence-error curve for semi-supervised learning with Stein Variational Gradient Descent obtained by considering both epistemic and aleatoric uncertainty is below (or matches) that obtained by considering epistemic or aleatoric uncertainty alone. Considering both sources of uncertainty leads to much more accurate predictions at the high confidence limit for $\mathtt{ESOL}$ and $\mathtt{p450}$. Moreover, there is no consistent trend as to whether epistemic or aleatoric uncertainty is more important -- for $\mathtt{ESOL}$, epistemic uncertainty is a better estimate of error than aleatoric uncertainty, whereas the opposite is true for $\mathtt{p450}$ and $\mathtt{CatS}$. As such, one cannot overlook epistemic or aleatoric uncertainty a priori, and our approach of combining both sources of uncertainty leads to an accurate uncertainty estimate.

 \begin{figure*}[!h]
\centering
\includegraphics[scale=0.17]{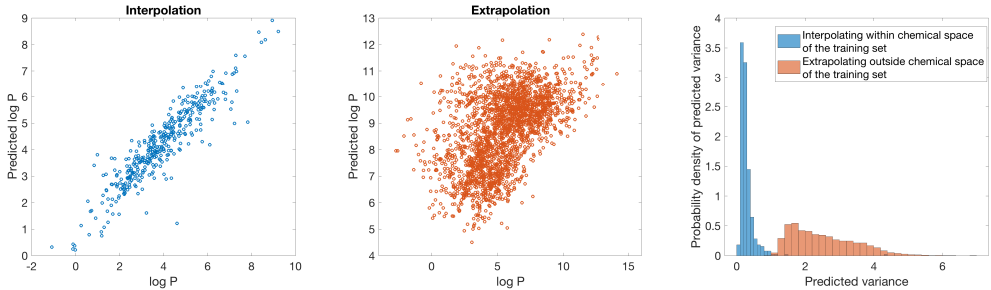}
\caption{Dataset bias can be mitigated with Bayesian uncertainty estimation. We consider a toy problem of predicting computationally calculated $\log P$ using Stein Variational Gradient Descent and semi-supervised learning, with a biased dataset comprising all beta-lactams or a benzodiazepines from ChEMBL. (Left) The model performance when the test set is also drawn from beta-lactams and benzodiazepines. (Middle) The model performance when the test set is all steroids from ChEMBL. (Right) The distribution of predicted uncertainty for the model applied to steroids and the model applied to beta-lactams and benzodiazepines. }
\label{biased} 
\end{figure*}  

\subsection{Overcoming dataset bias}

Our Bayesian methodology also overcomes dataset bias, which has be noted in the recent literature as the leading cause for overly optimistic results on benchmarks \cite{wallach2018most}. Most ligand-based benchmarks are biased in the sense that the molecules reported are tightly clustered around a few important chemical scaffolds, such that when the dataset is randomly split into training set and test set, the molecules in the test set are structurally very similar to the training set. Therefore, a model that only memorises the training set will still achieve a high accuracy on the test set yet cannot generalise to other regions of chemical space. Methods such as scaffold splitting \cite{wu2018moleculenet} and attribution \cite{mccloskey2018using} attempt to estimate what would be the true performance of the model if the dataset were not biased. However, bias is fundamental in chemical data -- an ''uniform distribution'' in chemical space does not exist because chemical space does not have a well-defined metric. Regardless of how one preprocesses the dataset or train the model, model predictions will always be awry for scaffolds that are not represented in the dataset. As such, rather than ``unbiasing'' the data, the practical question is whether the model can estimate whether it is likely to make a correct prediction for an unseen molecule given a biased training set. 

To show that Bayesian uncertainty estimation overcomes dataset bias, we consider a toy problem where we know the ground truth and deliberately introduce bias: We consider the problem of predicting octanol-water partition coefficient ($\log P$) values, and use computed $\mathrm{ACD} \log P$ values as a surrogate. We construct a dataset comprising all molecules on ChEMBL with either a beta-lactam or a benzodiazepine scaffold. The dataset is obviously very biased as it only contains 2 scaffolds. We then train a model using Stein Variational Gradient Descent with semi-supervised learning, with the standard 8:2 split between training and test set on the biased dataset. Figure \ref{biased} (left) show that model is reasonably accurate on the biased test set. We now simulate how an user might unwittingly fall foul of dataset bias -- suppose we use the model to predict $\log P$ of all molecules with a steroid scaffold on ChEMBL. Figure \ref{biased} (middle) show that the model performance, perhaps unsurprisingly, is poor. Steroids are not part of the training set, thus the model cannot predict its physiochemical properties. Bayesian uncertainty estimation provides a way out of this quandary -- Figure \ref{biased} (right) shows that the estimated uncertainty of $\log P$ prediction on steroids is significantly greater than $\log P$ prediction on the test set of beta-lactams or a benzodiazepines. In other words, the model can inform the user when it is inaccurate, thus mitigating the impact of dataset bias. 

\subsection{Low data active learning}

Having considered the quality of the uncertainty estimates in the data-abundant limit, our next question is whether we can estimate uncertainty in the low data limit and drive an active learning cycle. We consider the objective of obtaining a low model error with a small training set. The model is first trained from a small initial pool of data (25 \% of the full training set, picked randomly), the model then selects a batch of molecules (2.5\% of the full training set) that has the largest epistemic uncertainty to put into the training set, and then the model is retrained to suggest other additions, and the cycle continues. The test set is always 20\% of the full dataset, held out at the beginning of the experiment. We note that other acquisition functions have been suggested in the literature \cite{gal2017deep}, and the objective function is problem-dependent \cite{reker2015active,reker2017active}. Nonetheless, the goal of our experiment is to evaluate the quality of uncertainty estimate, thus we focus on a simple objective and acquisition functions. Further, as active learning requires constant retraining of the model, and Stein Variational Gradient Descent is significantly more computationally intensive than dropout, we will only consider dropout variational inference. 

\begin{figure*}[!h]
\centering
\includegraphics[scale=0.22]{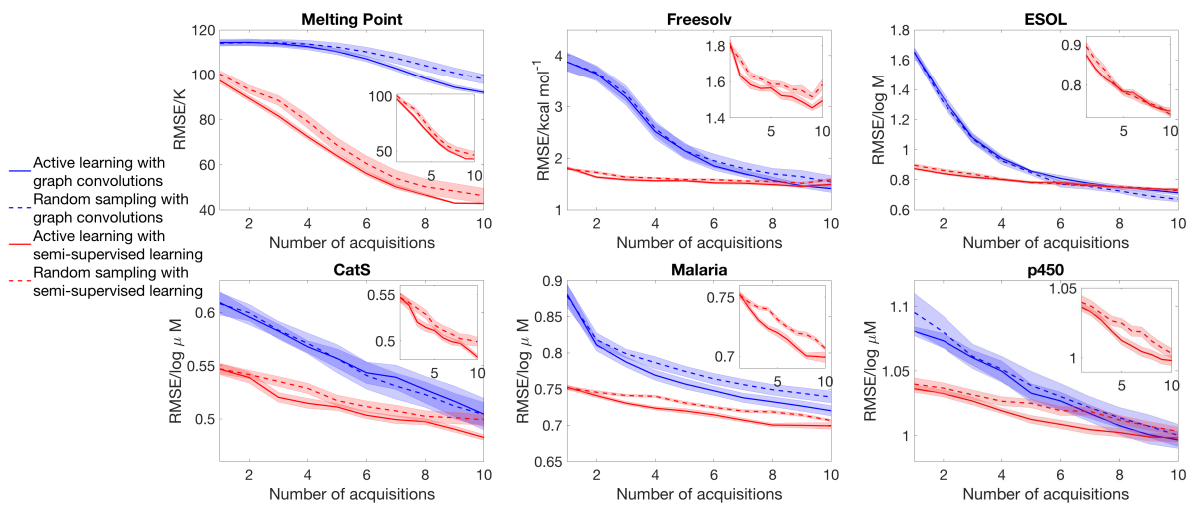}
\caption{Semi-supervised learning significantly outperforms full supervised learning in active learning. The model starts with 25 \% of the full training set, selected randomly, and at each iteration 2.5\% of the full training set is added to the training set. The molecules added are picked randomly (random sampling) or picked because they have the largest predicted epistemic uncertainty (active learning). The curves show the mean model error and standard error of the mean, averaged over 20 active learning runs starting from random dataset splits, as a function of iteration. The insets focus on the performance of semi-supervised learning with SVGD. }
\label{active_learning} 
\end{figure*}

Figure \ref{active_learning} shows that semi-supervised learning significantly outperforms full supervised learning in the low data limit. The mean learning curves and error bars are obtained by analysing 20 active learning runs starting from random dataset splits. Moreover, in the case of full supervised learning, active learning is unable to deliver a better learning curve than random sampling, whereas for semi-supervised learning there is a sizeable gap between the learning curves of random sampling and active learning. This is because the full supervised method generates molecular descriptors directly from data. Therefore, in the low data limit, it is unable to learn descriptors that describe the structure of chemical space and chemical similarity between compounds, thus cannot generate meaningful uncertainty estimates to drive active learning.

\begin{figure}[!h]
\centering
\includegraphics[scale=0.25]{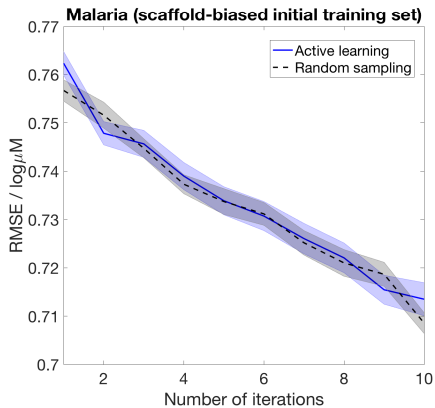}
\caption{Choosing diverse compounds in the initial screen is crucial to successful active learning. We first randomly split the $\mathtt{Malaria}$ dataset into training (80\%) and test (20\%) sets. We then scaffold-split the training set to obtain a biased initial set (25\% of the total training set), and at each iteration 2.5\% of the training set is given to the model, selected randomly (random sampling) or based on highest epistemic uncertainty (active learning).}
\label{malaria_scaffold} 
\end{figure}

The importance of choosing diverse compounds in the initial screen has been discussed extensively in the literature \cite{huggin2011,bakken2012,paricharak2018}, and the performance of our active learning method also depends on the chemical diversity in the initial screen. Figure \ref{malaria_scaffold} shows that active learning does not outperform random sampling when the initial training set biased and contain only a small number of scaffolds.  We model scaffold bias by splitting the data using scaffold splitting implemented in $\mathtt{DeepChem}$ \cite{wu2018moleculenet}, and consider the $\mathtt{Malaria}$ example where active learning most clearly outperforms random sampling in Figure \ref{active_learning}. The underperformance of active learning is perhaps unsurprising -- if the initial screen only consists of one scaffold, the knowledge that the model has on the other scaffolds would be minimal, i.e. the model is equally ignorant about all the other scaffolds. As such, randomly sampling the other scaffolds becomes a reasonable strategy.

\section{Discussion and Conclusion}
We propose a novel method to quantify uncertainty in molecular properties prediction. We show that our methodology significantly outperforms the baseline on a range of benchmark problems, both in terms of model accuracy and in terms of uncertainty estimates. Our method also overcomes dataset bias by returning a large uncertainty estimate when the test set is drawn from a different region of chemical space compared to the training set. Moreover, our methodology can drive an active learning cycle, maximising model performance while minimising the size of the training set. The key to the success of our method is the combination of semi-supervised learning and Bayesian deep learning. Semi-supervised learning allows us to learn informative molecular representation in the low data limit. Bayesian deep learning allows us to estimate aleatoric and epistemic uncertainty in a statistically principled manner. We exemplified our methodology on regression as it is generally more challenging than classification, although it can be readily extended to classification problems. 

Our observation that the choice of Bayesian inference methodology significantly impacts the quality of the uncertainty estimate suggests an evident followup that probes the mathematical limit of Bayesian inference -- i.e. benchmarking approximate inference techniques against importance sampling of the posterior using Markov Chain Monte Carlo till convergence, which is computationally expensive but mathematically exact. Moreover, we note that most approximate inference techniques in the literature have been benchmarked in terms of RMSE error or log-likelihood \cite{mukhoti2018importance}, rather than explicitly considering the quality of the uncertainty estimate in a manner relevant for chemoinformatics such as the confidence-error curve. An open question is the design of appropriate approximate inference techniques for graph convolutional neural networks that solves the trilemma between computational cost, model accuracy, and the quality of uncertainty estimate. 

Another open question is whether the model has accurately disentangled aleatoric and epistemic uncertainty. Answering this question would require estimates of the ground truth aleatoric uncertainty, which is obtainable via repeating the experimental measurement and reporting the variance. Benchmark datasets which provide accurate experimental uncertainty estimates will be invaluable to advancing the Bayesian methodology. 

Finally, our active learning methodology performs well when the initial screen covers diverse compounds. To successfully perform active learning on a scaffold-biased initial set, the model needs information on the bioactivity of those unseen scaffolds. We speculate that strategies such as multitask learning \cite{ramsundar2017,wenzel2019}, which pools information from other cognate assays which have explored the unseen scaffolds, will be a fruitful avenue.

\begin{acknowledgements}
AAL acknowledges the support of the Winton Programme for the Physics of Sustainability. 
\end{acknowledgements}

\bibliography{refs1} 

\begin{thebibliography}{47}%
\makeatletter
\providecommand \@ifxundefined [1]{%
 \@ifx{#1\undefined}
}%
\providecommand \@ifnum [1]{%
 \ifnum #1\expandafter \@firstoftwo
 \else \expandafter \@secondoftwo
 \fi
}%
\providecommand \@ifx [1]{%
 \ifx #1\expandafter \@firstoftwo
 \else \expandafter \@secondoftwo
 \fi
}%
\providecommand \natexlab [1]{#1}%
\providecommand \enquote  [1]{``#1''}%
\providecommand \bibnamefont  [1]{#1}%
\providecommand \bibfnamefont [1]{#1}%
\providecommand \citenamefont [1]{#1}%
\providecommand \href@noop [0]{\@secondoftwo}%
\providecommand \href [0]{\begingroup \@sanitize@url \@href}%
\providecommand \@href[1]{\@@startlink{#1}\@@href}%
\providecommand \@@href[1]{\endgroup#1\@@endlink}%
\providecommand \@sanitize@url [0]{\catcode `\\12\catcode `\$12\catcode
  `\&12\catcode `\#12\catcode `\^12\catcode `\_12\catcode `\%12\relax}%
\providecommand \@@startlink[1]{}%
\providecommand \@@endlink[0]{}%
\providecommand \url  [0]{\begingroup\@sanitize@url \@url }%
\providecommand \@url [1]{\endgroup\@href {#1}{\urlprefix }}%
\providecommand \urlprefix  [0]{URL }%
\providecommand \Eprint [0]{\href }%
\providecommand \doibase [0]{http://dx.doi.org/}%
\providecommand \selectlanguage [0]{\@gobble}%
\providecommand \bibinfo  [0]{\@secondoftwo}%
\providecommand \bibfield  [0]{\@secondoftwo}%
\providecommand \translation [1]{[#1]}%
\providecommand \BibitemOpen [0]{}%
\providecommand \bibitemStop [0]{}%
\providecommand \bibitemNoStop [0]{.\EOS\space}%
\providecommand \EOS [0]{\spacefactor3000\relax}%
\providecommand \BibitemShut  [1]{\csname bibitem#1\endcsname}%
\let\auto@bib@innerbib\@empty
\bibitem [{\citenamefont {Cherkasov}\ \emph {et~al.}(2014)\citenamefont
  {Cherkasov}, \citenamefont {Muratov}, \citenamefont {Fourches}, \citenamefont
  {Varnek}, \citenamefont {Baskin}, \citenamefont {Cronin}, \citenamefont
  {Dearden}, \citenamefont {Gramatica}, \citenamefont {Martin}, \citenamefont
  {Todeschini} \emph {et~al.}}]{cherkasov2014qsar}%
  \BibitemOpen
  \bibfield  {author} {\bibinfo {author} {\bibfnamefont {A.}~\bibnamefont
  {Cherkasov}}, \bibinfo {author} {\bibfnamefont {E.~N.}\ \bibnamefont
  {Muratov}}, \bibinfo {author} {\bibfnamefont {D.}~\bibnamefont {Fourches}},
  \bibinfo {author} {\bibfnamefont {A.}~\bibnamefont {Varnek}}, \bibinfo
  {author} {\bibfnamefont {I.~I.}\ \bibnamefont {Baskin}}, \bibinfo {author}
  {\bibfnamefont {M.}~\bibnamefont {Cronin}}, \bibinfo {author} {\bibfnamefont
  {J.}~\bibnamefont {Dearden}}, \bibinfo {author} {\bibfnamefont
  {P.}~\bibnamefont {Gramatica}}, \bibinfo {author} {\bibfnamefont {Y.~C.}\
  \bibnamefont {Martin}}, \bibinfo {author} {\bibfnamefont {R.}~\bibnamefont
  {Todeschini}},  \emph {et~al.},\ }\href@noop {} {\bibfield  {journal}
  {\bibinfo  {journal} {Journal of medicinal chemistry}\ }\textbf {\bibinfo
  {volume} {57}},\ \bibinfo {pages} {4977} (\bibinfo {year}
  {2014})}\BibitemShut {NoStop}%
\bibitem [{\citenamefont {Randi{\'c}}(1991)}]{randic1991generalized}%
  \BibitemOpen
  \bibfield  {author} {\bibinfo {author} {\bibfnamefont {M.}~\bibnamefont
  {Randi{\'c}}},\ }\href@noop {} {\bibfield  {journal} {\bibinfo  {journal}
  {Journal of Mathematical Chemistry}\ }\textbf {\bibinfo {volume} {7}},\
  \bibinfo {pages} {155} (\bibinfo {year} {1991})}\BibitemShut {NoStop}%
\bibitem [{\citenamefont {Ivanciuc}(2000)}]{ivanciuc2000qsar}%
  \BibitemOpen
  \bibfield  {author} {\bibinfo {author} {\bibfnamefont {O.}~\bibnamefont
  {Ivanciuc}},\ }\href@noop {} {\bibfield  {journal} {\bibinfo  {journal}
  {Journal of chemical information and computer sciences}\ }\textbf {\bibinfo
  {volume} {40}},\ \bibinfo {pages} {1412} (\bibinfo {year}
  {2000})}\BibitemShut {NoStop}%
\bibitem [{\citenamefont {Durant}\ \emph {et~al.}(2002)\citenamefont {Durant},
  \citenamefont {Leland}, \citenamefont {Henry},\ and\ \citenamefont
  {Nourse}}]{durant2002reoptimization}%
  \BibitemOpen
  \bibfield  {author} {\bibinfo {author} {\bibfnamefont {J.~L.}\ \bibnamefont
  {Durant}}, \bibinfo {author} {\bibfnamefont {B.~A.}\ \bibnamefont {Leland}},
  \bibinfo {author} {\bibfnamefont {D.~R.}\ \bibnamefont {Henry}}, \ and\
  \bibinfo {author} {\bibfnamefont {J.~G.}\ \bibnamefont {Nourse}},\
  }\href@noop {} {\bibfield  {journal} {\bibinfo  {journal} {Journal of
  chemical information and computer sciences}\ }\textbf {\bibinfo {volume}
  {42}},\ \bibinfo {pages} {1273} (\bibinfo {year} {2002})}\BibitemShut
  {NoStop}%
\bibitem [{\citenamefont {Rogers}\ \emph {et~al.}(2005)\citenamefont {Rogers},
  \citenamefont {Brown},\ and\ \citenamefont {Hahn}}]{rogers2005using}%
  \BibitemOpen
  \bibfield  {author} {\bibinfo {author} {\bibfnamefont {D.}~\bibnamefont
  {Rogers}}, \bibinfo {author} {\bibfnamefont {R.~D.}\ \bibnamefont {Brown}}, \
  and\ \bibinfo {author} {\bibfnamefont {M.}~\bibnamefont {Hahn}},\ }\href@noop
  {} {\bibfield  {journal} {\bibinfo  {journal} {Journal of biomolecular
  screening}\ }\textbf {\bibinfo {volume} {10}},\ \bibinfo {pages} {682}
  (\bibinfo {year} {2005})}\BibitemShut {NoStop}%
\bibitem [{\citenamefont {Cramer}\ \emph {et~al.}(1988)\citenamefont {Cramer},
  \citenamefont {Patterson},\ and\ \citenamefont
  {Bunce}}]{cramer1988comparative}%
  \BibitemOpen
  \bibfield  {author} {\bibinfo {author} {\bibfnamefont {R.~D.}\ \bibnamefont
  {Cramer}}, \bibinfo {author} {\bibfnamefont {D.~E.}\ \bibnamefont
  {Patterson}}, \ and\ \bibinfo {author} {\bibfnamefont {J.~D.}\ \bibnamefont
  {Bunce}},\ }\href@noop {} {\bibfield  {journal} {\bibinfo  {journal} {Journal
  of the American Chemical Society}\ }\textbf {\bibinfo {volume} {110}},\
  \bibinfo {pages} {5959} (\bibinfo {year} {1988})}\BibitemShut {NoStop}%
\bibitem [{\citenamefont {Verma}\ \emph {et~al.}(2010)\citenamefont {Verma},
  \citenamefont {Khedkar},\ and\ \citenamefont {Coutinho}}]{verma20103d}%
  \BibitemOpen
  \bibfield  {author} {\bibinfo {author} {\bibfnamefont {J.}~\bibnamefont
  {Verma}}, \bibinfo {author} {\bibfnamefont {V.~M.}\ \bibnamefont {Khedkar}},
  \ and\ \bibinfo {author} {\bibfnamefont {E.~C.}\ \bibnamefont {Coutinho}},\
  }\href@noop {} {\bibfield  {journal} {\bibinfo  {journal} {Current topics in
  medicinal chemistry}\ }\textbf {\bibinfo {volume} {10}},\ \bibinfo {pages}
  {95} (\bibinfo {year} {2010})}\BibitemShut {NoStop}%
\bibitem [{\citenamefont {Scarselli}\ \emph {et~al.}(2009)\citenamefont
  {Scarselli}, \citenamefont {Gori}, \citenamefont {Tsoi}, \citenamefont
  {Hagenbuchner},\ and\ \citenamefont {Monfardini}}]{scarselli2009graph}%
  \BibitemOpen
  \bibfield  {author} {\bibinfo {author} {\bibfnamefont {F.}~\bibnamefont
  {Scarselli}}, \bibinfo {author} {\bibfnamefont {M.}~\bibnamefont {Gori}},
  \bibinfo {author} {\bibfnamefont {A.~C.}\ \bibnamefont {Tsoi}}, \bibinfo
  {author} {\bibfnamefont {M.}~\bibnamefont {Hagenbuchner}}, \ and\ \bibinfo
  {author} {\bibfnamefont {G.}~\bibnamefont {Monfardini}},\ }\href@noop {}
  {\bibfield  {journal} {\bibinfo  {journal} {IEEE Transactions on Neural
  Networks}\ }\textbf {\bibinfo {volume} {20}},\ \bibinfo {pages} {61}
  (\bibinfo {year} {2009})}\BibitemShut {NoStop}%
\bibitem [{\citenamefont {Duvenaud}\ \emph {et~al.}(2015)\citenamefont
  {Duvenaud}, \citenamefont {Maclaurin}, \citenamefont {Iparraguirre},
  \citenamefont {Bombarell}, \citenamefont {Hirzel}, \citenamefont
  {Aspuru-Guzik},\ and\ \citenamefont {Adams}}]{duvenaud2015convolutional}%
  \BibitemOpen
  \bibfield  {author} {\bibinfo {author} {\bibfnamefont {D.~K.}\ \bibnamefont
  {Duvenaud}}, \bibinfo {author} {\bibfnamefont {D.}~\bibnamefont {Maclaurin}},
  \bibinfo {author} {\bibfnamefont {J.}~\bibnamefont {Iparraguirre}}, \bibinfo
  {author} {\bibfnamefont {R.}~\bibnamefont {Bombarell}}, \bibinfo {author}
  {\bibfnamefont {T.}~\bibnamefont {Hirzel}}, \bibinfo {author} {\bibfnamefont
  {A.}~\bibnamefont {Aspuru-Guzik}}, \ and\ \bibinfo {author} {\bibfnamefont
  {R.~P.}\ \bibnamefont {Adams}},\ }in\ \href@noop {} {\emph {\bibinfo
  {booktitle} {Advances in neural information processing systems}}}\ (\bibinfo
  {year} {2015})\ pp.\ \bibinfo {pages} {2224--2232}\BibitemShut {NoStop}%
\bibitem [{\citenamefont {Wu}\ \emph {et~al.}(2018)\citenamefont {Wu},
  \citenamefont {Ramsundar}, \citenamefont {Feinberg}, \citenamefont {Gomes},
  \citenamefont {Geniesse}, \citenamefont {Pappu}, \citenamefont {Leswing},\
  and\ \citenamefont {Pande}}]{wu2018moleculenet}%
  \BibitemOpen
  \bibfield  {author} {\bibinfo {author} {\bibfnamefont {Z.}~\bibnamefont
  {Wu}}, \bibinfo {author} {\bibfnamefont {B.}~\bibnamefont {Ramsundar}},
  \bibinfo {author} {\bibfnamefont {E.~N.}\ \bibnamefont {Feinberg}}, \bibinfo
  {author} {\bibfnamefont {J.}~\bibnamefont {Gomes}}, \bibinfo {author}
  {\bibfnamefont {C.}~\bibnamefont {Geniesse}}, \bibinfo {author}
  {\bibfnamefont {A.~S.}\ \bibnamefont {Pappu}}, \bibinfo {author}
  {\bibfnamefont {K.}~\bibnamefont {Leswing}}, \ and\ \bibinfo {author}
  {\bibfnamefont {V.}~\bibnamefont {Pande}},\ }\href@noop {} {\bibfield
  {journal} {\bibinfo  {journal} {Chemical science}\ }\textbf {\bibinfo
  {volume} {9}},\ \bibinfo {pages} {513} (\bibinfo {year} {2018})}\BibitemShut
  {NoStop}%
\bibitem [{\citenamefont {Sheridan}\ \emph {et~al.}(2004)\citenamefont
  {Sheridan}, \citenamefont {Feuston}, \citenamefont {Maiorov},\ and\
  \citenamefont {Kearsley}}]{sheridan2004similarity}%
  \BibitemOpen
  \bibfield  {author} {\bibinfo {author} {\bibfnamefont {R.~P.}\ \bibnamefont
  {Sheridan}}, \bibinfo {author} {\bibfnamefont {B.~P.}\ \bibnamefont
  {Feuston}}, \bibinfo {author} {\bibfnamefont {V.~N.}\ \bibnamefont
  {Maiorov}}, \ and\ \bibinfo {author} {\bibfnamefont {S.~K.}\ \bibnamefont
  {Kearsley}},\ }\href@noop {} {\bibfield  {journal} {\bibinfo  {journal}
  {Journal of Chemical Information and Computer Sciences}\ }\textbf {\bibinfo
  {volume} {44}},\ \bibinfo {pages} {1912} (\bibinfo {year}
  {2004})}\BibitemShut {NoStop}%
\bibitem [{\citenamefont {Sushko}\ \emph {et~al.}(2010)\citenamefont {Sushko},
  \citenamefont {Novotarskyi}, \citenamefont {Ko~rner}, \citenamefont {Pandey},
  \citenamefont {Cherkasov}, \citenamefont {Li}, \citenamefont {Gramatica},
  \citenamefont {Hansen}, \citenamefont {Schroeter}, \citenamefont {Mu~ller}
  \emph {et~al.}}]{sushko2010applicability}%
  \BibitemOpen
  \bibfield  {author} {\bibinfo {author} {\bibfnamefont {I.}~\bibnamefont
  {Sushko}}, \bibinfo {author} {\bibfnamefont {S.}~\bibnamefont {Novotarskyi}},
  \bibinfo {author} {\bibfnamefont {R.}~\bibnamefont {Ko~rner}}, \bibinfo
  {author} {\bibfnamefont {A.~K.}\ \bibnamefont {Pandey}}, \bibinfo {author}
  {\bibfnamefont {A.}~\bibnamefont {Cherkasov}}, \bibinfo {author}
  {\bibfnamefont {J.}~\bibnamefont {Li}}, \bibinfo {author} {\bibfnamefont
  {P.}~\bibnamefont {Gramatica}}, \bibinfo {author} {\bibfnamefont
  {K.}~\bibnamefont {Hansen}}, \bibinfo {author} {\bibfnamefont
  {T.}~\bibnamefont {Schroeter}}, \bibinfo {author} {\bibfnamefont {K.-R.}\
  \bibnamefont {Mu~ller}},  \emph {et~al.},\ }\href@noop {} {\bibfield
  {journal} {\bibinfo  {journal} {Journal of chemical information and
  modeling}\ }\textbf {\bibinfo {volume} {50}},\ \bibinfo {pages} {2094}
  (\bibinfo {year} {2010})}\BibitemShut {NoStop}%
\bibitem [{\citenamefont {Sheridan}(2012)}]{sheridan2012three}%
  \BibitemOpen
  \bibfield  {author} {\bibinfo {author} {\bibfnamefont {R.~P.}\ \bibnamefont
  {Sheridan}},\ }\href@noop {} {\bibfield  {journal} {\bibinfo  {journal}
  {Journal of chemical information and modeling}\ }\textbf {\bibinfo {volume}
  {52}},\ \bibinfo {pages} {814} (\bibinfo {year} {2012})}\BibitemShut
  {NoStop}%
\bibitem [{\citenamefont {Toplak}\ \emph {et~al.}(2014)\citenamefont {Toplak},
  \citenamefont {Moc~nik}, \citenamefont {Polajnar}, \citenamefont {Bosnic},
  \citenamefont {Carlsson}, \citenamefont {Hasselgren}, \citenamefont
  {Dems~ar}, \citenamefont {Boyer}, \citenamefont {Zupan},\ and\ \citenamefont
  {Sta~lring}}]{toplak2014assessment}%
  \BibitemOpen
  \bibfield  {author} {\bibinfo {author} {\bibfnamefont {M.}~\bibnamefont
  {Toplak}}, \bibinfo {author} {\bibfnamefont {R.}~\bibnamefont {Moc~nik}},
  \bibinfo {author} {\bibfnamefont {M.}~\bibnamefont {Polajnar}}, \bibinfo
  {author} {\bibfnamefont {Z.}~\bibnamefont {Bosnic}}, \bibinfo {author}
  {\bibfnamefont {L.}~\bibnamefont {Carlsson}}, \bibinfo {author}
  {\bibfnamefont {C.}~\bibnamefont {Hasselgren}}, \bibinfo {author}
  {\bibfnamefont {J.}~\bibnamefont {Dems~ar}}, \bibinfo {author} {\bibfnamefont
  {S.}~\bibnamefont {Boyer}}, \bibinfo {author} {\bibfnamefont
  {B.}~\bibnamefont {Zupan}}, \ and\ \bibinfo {author} {\bibfnamefont
  {J.}~\bibnamefont {Sta~lring}},\ }\href@noop {} {\bibfield  {journal}
  {\bibinfo  {journal} {Journal of chemical information and modeling}\ }\textbf
  {\bibinfo {volume} {54}},\ \bibinfo {pages} {431} (\bibinfo {year}
  {2014})}\BibitemShut {NoStop}%
\bibitem [{\citenamefont {Norinder}\ \emph {et~al.}(2014)\citenamefont
  {Norinder}, \citenamefont {Carlsson}, \citenamefont {Boyer},\ and\
  \citenamefont {Eklund}}]{norinder2014introducing}%
  \BibitemOpen
  \bibfield  {author} {\bibinfo {author} {\bibfnamefont {U.}~\bibnamefont
  {Norinder}}, \bibinfo {author} {\bibfnamefont {L.}~\bibnamefont {Carlsson}},
  \bibinfo {author} {\bibfnamefont {S.}~\bibnamefont {Boyer}}, \ and\ \bibinfo
  {author} {\bibfnamefont {M.}~\bibnamefont {Eklund}},\ }\href@noop {}
  {\bibfield  {journal} {\bibinfo  {journal} {Journal of chemical information
  and modeling}\ }\textbf {\bibinfo {volume} {54}},\ \bibinfo {pages} {1596}
  (\bibinfo {year} {2014})}\BibitemShut {NoStop}%
\bibitem [{\citenamefont {Svensson}\ \emph {et~al.}(2017)\citenamefont
  {Svensson}, \citenamefont {Norinder},\ and\ \citenamefont
  {Bender}}]{svensson2017modelling}%
  \BibitemOpen
  \bibfield  {author} {\bibinfo {author} {\bibfnamefont {F.}~\bibnamefont
  {Svensson}}, \bibinfo {author} {\bibfnamefont {U.}~\bibnamefont {Norinder}},
  \ and\ \bibinfo {author} {\bibfnamefont {A.}~\bibnamefont {Bender}},\
  }\href@noop {} {\bibfield  {journal} {\bibinfo  {journal} {Toxicology
  Research}\ }\textbf {\bibinfo {volume} {6}},\ \bibinfo {pages} {73} (\bibinfo
  {year} {2017})}\BibitemShut {NoStop}%
\bibitem [{\citenamefont {Kendall}\ and\ \citenamefont
  {Gal}(2017)}]{kendall2017uncertainties}%
  \BibitemOpen
  \bibfield  {author} {\bibinfo {author} {\bibfnamefont {A.}~\bibnamefont
  {Kendall}}\ and\ \bibinfo {author} {\bibfnamefont {Y.}~\bibnamefont {Gal}},\
  }in\ \href@noop {} {\emph {\bibinfo {booktitle} {Advances in neural
  information processing systems}}}\ (\bibinfo {year} {2017})\ pp.\ \bibinfo
  {pages} {5574--5584}\BibitemShut {NoStop}%
\bibitem [{\citenamefont {Reker}\ and\ \citenamefont
  {Schneider}(2015)}]{reker2015active}%
  \BibitemOpen
  \bibfield  {author} {\bibinfo {author} {\bibfnamefont {D.}~\bibnamefont
  {Reker}}\ and\ \bibinfo {author} {\bibfnamefont {G.}~\bibnamefont
  {Schneider}},\ }\href@noop {} {\bibfield  {journal} {\bibinfo  {journal}
  {Drug discovery today}\ }\textbf {\bibinfo {volume} {20}},\ \bibinfo {pages}
  {458} (\bibinfo {year} {2015})}\BibitemShut {NoStop}%
\bibitem [{\citenamefont {Reker}\ \emph {et~al.}(2017)\citenamefont {Reker},
  \citenamefont {Schneider}, \citenamefont {Schneider},\ and\ \citenamefont
  {Brown}}]{reker2017active}%
  \BibitemOpen
  \bibfield  {author} {\bibinfo {author} {\bibfnamefont {D.}~\bibnamefont
  {Reker}}, \bibinfo {author} {\bibfnamefont {P.}~\bibnamefont {Schneider}},
  \bibinfo {author} {\bibfnamefont {G.}~\bibnamefont {Schneider}}, \ and\
  \bibinfo {author} {\bibfnamefont {J.}~\bibnamefont {Brown}},\ }\href@noop {}
  {\bibfield  {journal} {\bibinfo  {journal} {Future medicinal chemistry}\
  }\textbf {\bibinfo {volume} {9}},\ \bibinfo {pages} {381} (\bibinfo {year}
  {2017})}\BibitemShut {NoStop}%
\bibitem [{\citenamefont {Altae-Tran}\ \emph {et~al.}(2017)\citenamefont
  {Altae-Tran}, \citenamefont {Ramsundar}, \citenamefont {Pappu},\ and\
  \citenamefont {Pande}}]{altae2017low}%
  \BibitemOpen
  \bibfield  {author} {\bibinfo {author} {\bibfnamefont {H.}~\bibnamefont
  {Altae-Tran}}, \bibinfo {author} {\bibfnamefont {B.}~\bibnamefont
  {Ramsundar}}, \bibinfo {author} {\bibfnamefont {A.~S.}\ \bibnamefont
  {Pappu}}, \ and\ \bibinfo {author} {\bibfnamefont {V.}~\bibnamefont
  {Pande}},\ }\href@noop {} {\bibfield  {journal} {\bibinfo  {journal} {ACS
  central science}\ }\textbf {\bibinfo {volume} {3}},\ \bibinfo {pages} {283}
  (\bibinfo {year} {2017})}\BibitemShut {NoStop}%
\bibitem [{\citenamefont {G{\'o}mez-Bombarelli}\ \emph
  {et~al.}(2018)\citenamefont {G{\'o}mez-Bombarelli}, \citenamefont {Wei},
  \citenamefont {Duvenaud}, \citenamefont {Hern{\'a}ndez-Lobato}, \citenamefont
  {S{\'a}nchez-Lengeling}, \citenamefont {Sheberla}, \citenamefont
  {Aguilera-Iparraguirre}, \citenamefont {Hirzel}, \citenamefont {Adams},\ and\
  \citenamefont {Aspuru-Guzik}}]{gomez2018automatic}%
  \BibitemOpen
  \bibfield  {author} {\bibinfo {author} {\bibfnamefont {R.}~\bibnamefont
  {G{\'o}mez-Bombarelli}}, \bibinfo {author} {\bibfnamefont {J.~N.}\
  \bibnamefont {Wei}}, \bibinfo {author} {\bibfnamefont {D.}~\bibnamefont
  {Duvenaud}}, \bibinfo {author} {\bibfnamefont {J.~M.}\ \bibnamefont
  {Hern{\'a}ndez-Lobato}}, \bibinfo {author} {\bibfnamefont {B.}~\bibnamefont
  {S{\'a}nchez-Lengeling}}, \bibinfo {author} {\bibfnamefont {D.}~\bibnamefont
  {Sheberla}}, \bibinfo {author} {\bibfnamefont {J.}~\bibnamefont
  {Aguilera-Iparraguirre}}, \bibinfo {author} {\bibfnamefont {T.~D.}\
  \bibnamefont {Hirzel}}, \bibinfo {author} {\bibfnamefont {R.~P.}\
  \bibnamefont {Adams}}, \ and\ \bibinfo {author} {\bibfnamefont
  {A.}~\bibnamefont {Aspuru-Guzik}},\ }\href@noop {} {\bibfield  {journal}
  {\bibinfo  {journal} {ACS central science}\ }\textbf {\bibinfo {volume}
  {4}},\ \bibinfo {pages} {268} (\bibinfo {year} {2018})}\BibitemShut {NoStop}%
\bibitem [{\citenamefont {Gilmer}\ \emph {et~al.}(2017)\citenamefont {Gilmer},
  \citenamefont {Schoenholz}, \citenamefont {Riley}, \citenamefont {Vinyals},\
  and\ \citenamefont {Dahl}}]{gilmer2017neural}%
  \BibitemOpen
  \bibfield  {author} {\bibinfo {author} {\bibfnamefont {J.}~\bibnamefont
  {Gilmer}}, \bibinfo {author} {\bibfnamefont {S.~S.}\ \bibnamefont
  {Schoenholz}}, \bibinfo {author} {\bibfnamefont {P.~F.}\ \bibnamefont
  {Riley}}, \bibinfo {author} {\bibfnamefont {O.}~\bibnamefont {Vinyals}}, \
  and\ \bibinfo {author} {\bibfnamefont {G.~E.}\ \bibnamefont {Dahl}},\
  }\href@noop {} {\bibfield  {journal} {\bibinfo  {journal} {arXiv preprint
  arXiv:1704.01212}\ } (\bibinfo {year} {2017})}\BibitemShut {NoStop}%
\bibitem [{Note1()}]{Note1}%
  \BibitemOpen
  \bibinfo {note} {$\protect \mathtt
  {https://github.com/debbiemarkslab/neural-fingerprint-theano}$}\BibitemShut
  {NoStop}%
\bibitem [{\citenamefont {Welsch}\ \emph {et~al.}(2010)\citenamefont {Welsch},
  \citenamefont {Snyder},\ and\ \citenamefont
  {Stockwell}}]{welsch2010privileged}%
  \BibitemOpen
  \bibfield  {author} {\bibinfo {author} {\bibfnamefont {M.~E.}\ \bibnamefont
  {Welsch}}, \bibinfo {author} {\bibfnamefont {S.~A.}\ \bibnamefont {Snyder}},
  \ and\ \bibinfo {author} {\bibfnamefont {B.~R.}\ \bibnamefont {Stockwell}},\
  }\href@noop {} {\bibfield  {journal} {\bibinfo  {journal} {Current opinion in
  chemical biology}\ }\textbf {\bibinfo {volume} {14}},\ \bibinfo {pages} {347}
  (\bibinfo {year} {2010})}\BibitemShut {NoStop}%
\bibitem [{\citenamefont {Nguyen}\ \emph {et~al.}(2017)\citenamefont {Nguyen},
  \citenamefont {Maeda},\ and\ \citenamefont {Oono}}]{nguyen2017semi}%
  \BibitemOpen
  \bibfield  {author} {\bibinfo {author} {\bibfnamefont {H.}~\bibnamefont
  {Nguyen}}, \bibinfo {author} {\bibfnamefont {S.-i.}\ \bibnamefont {Maeda}}, \
  and\ \bibinfo {author} {\bibfnamefont {K.}~\bibnamefont {Oono}},\ }\href@noop
  {} {\bibfield  {journal} {\bibinfo  {journal} {arXiv preprint
  arXiv:1711.10168}\ } (\bibinfo {year} {2017})}\BibitemShut {NoStop}%
\bibitem [{\citenamefont {Le}\ and\ \citenamefont
  {Mikolov}(2014)}]{le2014distributed}%
  \BibitemOpen
  \bibfield  {author} {\bibinfo {author} {\bibfnamefont {Q.}~\bibnamefont
  {Le}}\ and\ \bibinfo {author} {\bibfnamefont {T.}~\bibnamefont {Mikolov}},\
  }in\ \href@noop {} {\emph {\bibinfo {booktitle} {International Conference on
  Machine Learning}}}\ (\bibinfo {year} {2014})\ pp.\ \bibinfo {pages}
  {1188--1196}\BibitemShut {NoStop}%
\bibitem [{Note2()}]{Note2}%
  \BibitemOpen
  \bibinfo {note} {$\protect \mathtt
  {https://github.com/pfnet-research/hierarchical-molecular-learning}$}\BibitemShut
  {NoStop}%
\bibitem [{\citenamefont {Gal}\ and\ \citenamefont
  {Ghahramani}(2015)}]{gal2015dropout}%
  \BibitemOpen
  \bibfield  {author} {\bibinfo {author} {\bibfnamefont {Y.}~\bibnamefont
  {Gal}}\ and\ \bibinfo {author} {\bibfnamefont {Z.}~\bibnamefont
  {Ghahramani}},\ }\href@noop {} {\bibfield  {journal} {\bibinfo  {journal}
  {arXiv preprint arXiv:1506.02157}\ } (\bibinfo {year} {2015})}\BibitemShut
  {NoStop}%
\bibitem [{\citenamefont {Liu}\ and\ \citenamefont
  {Wang}(2016)}]{liu2016stein}%
  \BibitemOpen
  \bibfield  {author} {\bibinfo {author} {\bibfnamefont {Q.}~\bibnamefont
  {Liu}}\ and\ \bibinfo {author} {\bibfnamefont {D.}~\bibnamefont {Wang}},\
  }in\ \href@noop {} {\emph {\bibinfo {booktitle} {Advances In Neural
  Information Processing Systems}}}\ (\bibinfo {year} {2016})\ pp.\ \bibinfo
  {pages} {2378--2386}\BibitemShut {NoStop}%
\bibitem [{\citenamefont {Coley}\ \emph {et~al.}(2017)\citenamefont {Coley},
  \citenamefont {Barzilay}, \citenamefont {Green}, \citenamefont {Jaakkola},\
  and\ \citenamefont {Jensen}}]{coley2017convolutional}%
  \BibitemOpen
  \bibfield  {author} {\bibinfo {author} {\bibfnamefont {C.~W.}\ \bibnamefont
  {Coley}}, \bibinfo {author} {\bibfnamefont {R.}~\bibnamefont {Barzilay}},
  \bibinfo {author} {\bibfnamefont {W.~H.}\ \bibnamefont {Green}}, \bibinfo
  {author} {\bibfnamefont {T.~S.}\ \bibnamefont {Jaakkola}}, \ and\ \bibinfo
  {author} {\bibfnamefont {K.~F.}\ \bibnamefont {Jensen}},\ }\href@noop {}
  {\bibfield  {journal} {\bibinfo  {journal} {Journal of chemical information
  and modeling}\ }\textbf {\bibinfo {volume} {57}},\ \bibinfo {pages} {1757}
  (\bibinfo {year} {2017})}\BibitemShut {NoStop}%
\bibitem [{\citenamefont {Delaney}(2004)}]{delaney2004esol}%
  \BibitemOpen
  \bibfield  {author} {\bibinfo {author} {\bibfnamefont {J.~S.}\ \bibnamefont
  {Delaney}},\ }\href@noop {} {\bibfield  {journal} {\bibinfo  {journal}
  {Journal of chemical information and computer sciences}\ }\textbf {\bibinfo
  {volume} {44}},\ \bibinfo {pages} {1000} (\bibinfo {year}
  {2004})}\BibitemShut {NoStop}%
\bibitem [{\citenamefont {Mobley}\ and\ \citenamefont
  {Guthrie}(2014)}]{mobley2014freesolv}%
  \BibitemOpen
  \bibfield  {author} {\bibinfo {author} {\bibfnamefont {D.~L.}\ \bibnamefont
  {Mobley}}\ and\ \bibinfo {author} {\bibfnamefont {J.~P.}\ \bibnamefont
  {Guthrie}},\ }\href@noop {} {\bibfield  {journal} {\bibinfo  {journal}
  {Journal of computer-aided molecular design}\ }\textbf {\bibinfo {volume}
  {28}},\ \bibinfo {pages} {711} (\bibinfo {year} {2014})}\BibitemShut
  {NoStop}%
\bibitem [{\citenamefont {Gaieb}\ \emph {et~al.}(2019)\citenamefont {Gaieb},
  \citenamefont {Parks}, \citenamefont {Chiu}, \citenamefont {Yang},
  \citenamefont {Shao}, \citenamefont {Walters}, \citenamefont {Lambert},
  \citenamefont {Nevins}, \citenamefont {Bembenek}, \citenamefont {Ameriks},
  \citenamefont {Mirzadegan}, \citenamefont {Burley}, \citenamefont {Amaro},\
  and\ \citenamefont {Gilson}}]{Gaieb2019}%
  \BibitemOpen
  \bibfield  {author} {\bibinfo {author} {\bibfnamefont {Z.}~\bibnamefont
  {Gaieb}}, \bibinfo {author} {\bibfnamefont {C.~D.}\ \bibnamefont {Parks}},
  \bibinfo {author} {\bibfnamefont {M.}~\bibnamefont {Chiu}}, \bibinfo {author}
  {\bibfnamefont {H.}~\bibnamefont {Yang}}, \bibinfo {author} {\bibfnamefont
  {C.}~\bibnamefont {Shao}}, \bibinfo {author} {\bibfnamefont {W.~P.}\
  \bibnamefont {Walters}}, \bibinfo {author} {\bibfnamefont {M.~H.}\
  \bibnamefont {Lambert}}, \bibinfo {author} {\bibfnamefont {N.}~\bibnamefont
  {Nevins}}, \bibinfo {author} {\bibfnamefont {S.~D.}\ \bibnamefont
  {Bembenek}}, \bibinfo {author} {\bibfnamefont {M.~K.}\ \bibnamefont
  {Ameriks}}, \bibinfo {author} {\bibfnamefont {T.}~\bibnamefont {Mirzadegan}},
  \bibinfo {author} {\bibfnamefont {S.~K.}\ \bibnamefont {Burley}}, \bibinfo
  {author} {\bibfnamefont {R.~E.}\ \bibnamefont {Amaro}}, \ and\ \bibinfo
  {author} {\bibfnamefont {M.~K.}\ \bibnamefont {Gilson}},\ }\href@noop {}
  {\bibfield  {journal} {\bibinfo  {journal} {Journal of Computer-Aided
  Molecular Design}\ } (\bibinfo {year} {2019})}\BibitemShut {NoStop}%
\bibitem [{\citenamefont {Gamo}\ \emph {et~al.}(2010)\citenamefont {Gamo},
  \citenamefont {Sanz}, \citenamefont {Vidal}, \citenamefont {de~Cozar},
  \citenamefont {Alvarez}, \citenamefont {Lavandera}, \citenamefont
  {Vanderwall}, \citenamefont {Green}, \citenamefont {Kumar}, \citenamefont
  {Hasan} \emph {et~al.}}]{gamo2010thousands}%
  \BibitemOpen
  \bibfield  {author} {\bibinfo {author} {\bibfnamefont {F.-J.}\ \bibnamefont
  {Gamo}}, \bibinfo {author} {\bibfnamefont {L.~M.}\ \bibnamefont {Sanz}},
  \bibinfo {author} {\bibfnamefont {J.}~\bibnamefont {Vidal}}, \bibinfo
  {author} {\bibfnamefont {C.}~\bibnamefont {de~Cozar}}, \bibinfo {author}
  {\bibfnamefont {E.}~\bibnamefont {Alvarez}}, \bibinfo {author} {\bibfnamefont
  {J.-L.}\ \bibnamefont {Lavandera}}, \bibinfo {author} {\bibfnamefont {D.~E.}\
  \bibnamefont {Vanderwall}}, \bibinfo {author} {\bibfnamefont {D.~V.}\
  \bibnamefont {Green}}, \bibinfo {author} {\bibfnamefont {V.}~\bibnamefont
  {Kumar}}, \bibinfo {author} {\bibfnamefont {S.}~\bibnamefont {Hasan}},  \emph
  {et~al.},\ }\href@noop {} {\bibfield  {journal} {\bibinfo  {journal}
  {Nature}\ }\textbf {\bibinfo {volume} {465}},\ \bibinfo {pages} {305}
  (\bibinfo {year} {2010})}\BibitemShut {NoStop}%
\bibitem [{\citenamefont {Chen}\ and\ \citenamefont
  {Guestrin}(2016)}]{chen2016xgboost}%
  \BibitemOpen
  \bibfield  {author} {\bibinfo {author} {\bibfnamefont {T.}~\bibnamefont
  {Chen}}\ and\ \bibinfo {author} {\bibfnamefont {C.}~\bibnamefont
  {Guestrin}},\ }in\ \href@noop {} {\emph {\bibinfo {booktitle} {Proceedings of
  the 22nd acm sigkdd international conference on knowledge discovery and data
  mining}}}\ (\bibinfo {organization} {ACM},\ \bibinfo {year} {2016})\ pp.\
  \bibinfo {pages} {785--794}\BibitemShut {NoStop}%
\bibitem [{\citenamefont {Rogers}\ and\ \citenamefont
  {Hahn}(2010)}]{rogers2010}%
  \BibitemOpen
  \bibfield  {author} {\bibinfo {author} {\bibfnamefont {D.}~\bibnamefont
  {Rogers}}\ and\ \bibinfo {author} {\bibfnamefont {M.}~\bibnamefont {Hahn}},\
  }\href@noop {} {\bibfield  {journal} {\bibinfo  {journal} {Journal of
  Chemical Information and Modeling}\ }\textbf {\bibinfo {volume} {50}},\
  \bibinfo {pages} {742} (\bibinfo {year} {2010})}\BibitemShut {NoStop}%
\bibitem [{\citenamefont {Bender}\ and\ \citenamefont
  {Glen}(2005)}]{bender2005discussion}%
  \BibitemOpen
  \bibfield  {author} {\bibinfo {author} {\bibfnamefont {A.}~\bibnamefont
  {Bender}}\ and\ \bibinfo {author} {\bibfnamefont {R.~C.}\ \bibnamefont
  {Glen}},\ }\href@noop {} {\bibfield  {journal} {\bibinfo  {journal} {Journal
  of chemical information and modeling}\ }\textbf {\bibinfo {volume} {45}},\
  \bibinfo {pages} {1369} (\bibinfo {year} {2005})}\BibitemShut {NoStop}%
\bibitem [{\citenamefont {Truchon}\ and\ \citenamefont
  {Bayly}(2007)}]{truchon2007evaluating}%
  \BibitemOpen
  \bibfield  {author} {\bibinfo {author} {\bibfnamefont {J.-F.}\ \bibnamefont
  {Truchon}}\ and\ \bibinfo {author} {\bibfnamefont {C.~I.}\ \bibnamefont
  {Bayly}},\ }\href@noop {} {\bibfield  {journal} {\bibinfo  {journal} {Journal
  of chemical information and modeling}\ }\textbf {\bibinfo {volume} {47}},\
  \bibinfo {pages} {488} (\bibinfo {year} {2007})}\BibitemShut {NoStop}%
\bibitem [{\citenamefont {Wallach}\ and\ \citenamefont
  {Heifets}(2018)}]{wallach2018most}%
  \BibitemOpen
  \bibfield  {author} {\bibinfo {author} {\bibfnamefont {I.}~\bibnamefont
  {Wallach}}\ and\ \bibinfo {author} {\bibfnamefont {A.}~\bibnamefont
  {Heifets}},\ }\href@noop {} {\bibfield  {journal} {\bibinfo  {journal}
  {Journal of chemical information and modeling}\ }\textbf {\bibinfo {volume}
  {58}},\ \bibinfo {pages} {916} (\bibinfo {year} {2018})}\BibitemShut
  {NoStop}%
\bibitem [{\citenamefont {McCloskey}\ \emph {et~al.}(2018)\citenamefont
  {McCloskey}, \citenamefont {Taly}, \citenamefont {Monti}, \citenamefont
  {Brenner},\ and\ \citenamefont {Colwell}}]{mccloskey2018using}%
  \BibitemOpen
  \bibfield  {author} {\bibinfo {author} {\bibfnamefont {K.}~\bibnamefont
  {McCloskey}}, \bibinfo {author} {\bibfnamefont {A.}~\bibnamefont {Taly}},
  \bibinfo {author} {\bibfnamefont {F.}~\bibnamefont {Monti}}, \bibinfo
  {author} {\bibfnamefont {M.~P.}\ \bibnamefont {Brenner}}, \ and\ \bibinfo
  {author} {\bibfnamefont {L.}~\bibnamefont {Colwell}},\ }\href@noop {}
  {\bibfield  {journal} {\bibinfo  {journal} {arXiv preprint arXiv:1811.11310}\
  } (\bibinfo {year} {2018})}\BibitemShut {NoStop}%
\bibitem [{\citenamefont {Gal}\ \emph {et~al.}(2017)\citenamefont {Gal},
  \citenamefont {Islam},\ and\ \citenamefont {Ghahramani}}]{gal2017deep}%
  \BibitemOpen
  \bibfield  {author} {\bibinfo {author} {\bibfnamefont {Y.}~\bibnamefont
  {Gal}}, \bibinfo {author} {\bibfnamefont {R.}~\bibnamefont {Islam}}, \ and\
  \bibinfo {author} {\bibfnamefont {Z.}~\bibnamefont {Ghahramani}},\
  }\href@noop {} {\bibfield  {journal} {\bibinfo  {journal} {arXiv preprint
  arXiv:1703.02910}\ } (\bibinfo {year} {2017})}\BibitemShut {NoStop}%
\bibitem [{\citenamefont {Huggins}\ \emph {et~al.}(2011)\citenamefont
  {Huggins}, \citenamefont {Venkitaraman},\ and\ \citenamefont
  {Spring}}]{huggin2011}%
  \BibitemOpen
  \bibfield  {author} {\bibinfo {author} {\bibfnamefont {D.~J.}\ \bibnamefont
  {Huggins}}, \bibinfo {author} {\bibfnamefont {A.~R.}\ \bibnamefont
  {Venkitaraman}}, \ and\ \bibinfo {author} {\bibfnamefont {D.~R.}\
  \bibnamefont {Spring}},\ }\href@noop {} {\bibfield  {journal} {\bibinfo
  {journal} {ACS Chemical Biology}\ }\textbf {\bibinfo {volume} {6}},\ \bibinfo
  {pages} {208} (\bibinfo {year} {2011})}\BibitemShut {NoStop}%
\bibitem [{\citenamefont {Bakken}\ \emph {et~al.}(2012)\citenamefont {Bakken},
  \citenamefont {Bell}, \citenamefont {Boehm}, \citenamefont {Everett},
  \citenamefont {Gonzales}, \citenamefont {Hepworth}, \citenamefont
  {Klug-McLeod}, \citenamefont {Lanfear}, \citenamefont {Loesel}, \citenamefont
  {Mathias},\ and\ \citenamefont {Wood}}]{bakken2012}%
  \BibitemOpen
  \bibfield  {author} {\bibinfo {author} {\bibfnamefont {G.~A.}\ \bibnamefont
  {Bakken}}, \bibinfo {author} {\bibfnamefont {A.~S.}\ \bibnamefont {Bell}},
  \bibinfo {author} {\bibfnamefont {M.}~\bibnamefont {Boehm}}, \bibinfo
  {author} {\bibfnamefont {J.~R.}\ \bibnamefont {Everett}}, \bibinfo {author}
  {\bibfnamefont {R.}~\bibnamefont {Gonzales}}, \bibinfo {author}
  {\bibfnamefont {D.}~\bibnamefont {Hepworth}}, \bibinfo {author}
  {\bibfnamefont {J.~L.}\ \bibnamefont {Klug-McLeod}}, \bibinfo {author}
  {\bibfnamefont {J.}~\bibnamefont {Lanfear}}, \bibinfo {author} {\bibfnamefont
  {J.}~\bibnamefont {Loesel}}, \bibinfo {author} {\bibfnamefont
  {J.}~\bibnamefont {Mathias}}, \ and\ \bibinfo {author} {\bibfnamefont
  {T.~P.}\ \bibnamefont {Wood}},\ }\href@noop {} {\bibfield  {journal}
  {\bibinfo  {journal} {Journal of Chemical Information and Modeling}\ }\textbf
  {\bibinfo {volume} {52}},\ \bibinfo {pages} {2937} (\bibinfo {year}
  {2012})}\BibitemShut {NoStop}%
\bibitem [{\citenamefont {Paricharak}\ \emph {et~al.}(2018)\citenamefont
  {Paricharak}, \citenamefont {Mendez-Lucio}, \citenamefont {Ravindranath},
  \citenamefont {Bender}, \citenamefont {IJzerman},\ and\ \citenamefont {van
  Westen}}]{paricharak2018}%
  \BibitemOpen
  \bibfield  {author} {\bibinfo {author} {\bibfnamefont {S.}~\bibnamefont
  {Paricharak}}, \bibinfo {author} {\bibfnamefont {O.}~\bibnamefont
  {Mendez-Lucio}}, \bibinfo {author} {\bibfnamefont {A.~C.}\ \bibnamefont
  {Ravindranath}}, \bibinfo {author} {\bibfnamefont {A.}~\bibnamefont
  {Bender}}, \bibinfo {author} {\bibfnamefont {A.~P.}\ \bibnamefont
  {IJzerman}}, \ and\ \bibinfo {author} {\bibfnamefont {G.~J.~P.}\ \bibnamefont
  {van Westen}},\ }\href@noop {} {\bibfield  {journal} {\bibinfo  {journal}
  {Briefings in Bioinformatics}\ }\textbf {\bibinfo {volume} {19}},\ \bibinfo
  {pages} {277} (\bibinfo {year} {2018})}\BibitemShut {NoStop}%
\bibitem [{\citenamefont {Mukhoti}\ \emph {et~al.}(2018)\citenamefont
  {Mukhoti}, \citenamefont {Stenetorp},\ and\ \citenamefont
  {Gal}}]{mukhoti2018importance}%
  \BibitemOpen
  \bibfield  {author} {\bibinfo {author} {\bibfnamefont {J.}~\bibnamefont
  {Mukhoti}}, \bibinfo {author} {\bibfnamefont {P.}~\bibnamefont {Stenetorp}},
  \ and\ \bibinfo {author} {\bibfnamefont {Y.}~\bibnamefont {Gal}},\
  }\href@noop {} {\bibfield  {journal} {\bibinfo  {journal} {arXiv preprint
  arXiv:1811.09385}\ } (\bibinfo {year} {2018})}\BibitemShut {NoStop}%
\bibitem [{\citenamefont {Ramsunda}\ \emph {et~al.}(2017)\citenamefont
  {Ramsunda}, \citenamefont {Liu}, \citenamefont {Wu}, \citenamefont {Verras},
  \citenamefont {Tudor}, \citenamefont {Sheridan},\ and\ \citenamefont
  {Pande}}]{ramsundar2017}%
  \BibitemOpen
  \bibfield  {author} {\bibinfo {author} {\bibfnamefont {B.}~\bibnamefont
  {Ramsunda}}, \bibinfo {author} {\bibfnamefont {B.}~\bibnamefont {Liu}},
  \bibinfo {author} {\bibfnamefont {Z.}~\bibnamefont {Wu}}, \bibinfo {author}
  {\bibfnamefont {A.}~\bibnamefont {Verras}}, \bibinfo {author} {\bibfnamefont
  {M.}~\bibnamefont {Tudor}}, \bibinfo {author} {\bibfnamefont {R.~P.}\
  \bibnamefont {Sheridan}}, \ and\ \bibinfo {author} {\bibfnamefont
  {V.}~\bibnamefont {Pande}},\ }\href@noop {} {\bibfield  {journal} {\bibinfo
  {journal} {Journal of Chemical Information and Modeling}\ }\textbf {\bibinfo
  {volume} {57}},\ \bibinfo {pages} {2068} (\bibinfo {year}
  {2017})}\BibitemShut {NoStop}%
\bibitem [{\citenamefont {Wenzel}\ \emph {et~al.}(2019)\citenamefont {Wenzel},
  \citenamefont {Matter},\ and\ \citenamefont {Schmidt}}]{wenzel2019}%
  \BibitemOpen
  \bibfield  {author} {\bibinfo {author} {\bibfnamefont {J.}~\bibnamefont
  {Wenzel}}, \bibinfo {author} {\bibfnamefont {H.}~\bibnamefont {Matter}}, \
  and\ \bibinfo {author} {\bibfnamefont {F.}~\bibnamefont {Schmidt}},\
  }\href@noop {} {\bibfield  {journal} {\bibinfo  {journal} {Journal of
  Chemical Information and Modeling}\ }\textbf {\bibinfo {volume} {59}},\
  \bibinfo {pages} {1253} (\bibinfo {year} {2019})}\BibitemShut {NoStop}%
\end{thebibliography}%

\end{document}